\documentclass[letterpaper, 10 pt, conference]{ieeeconf}  % Comment this line out if you need a4paper

\IEEEoverridecommandlockouts                              % This command is only needed if 
\usepackage{caption}
\usepackage{booktabs}
\usepackage{tabularx}
\usepackage{graphics} % for pdf, bitmapped graphics files
\usepackage{epsfig} % for postscript graphics files
\usepackage{mathptmx} % assumes new font selection scheme installed
\usepackage{times} % assumes new font selection scheme installed
\usepackage{amsmath} % assumes amsmath package installed
\usepackage{amssymb}  % assumes amsmath package installed
\usepackage{xcolor}
\usepackage{amssymb}
\usepackage{hyperref}
\usepackage{algorithm}
\usepackage{algorithmic}
\newcommand{\redcross}{\textcolor{red}{\(\times\)}}
\newcommand{\greentick}{\textcolor{green}{\(\checkmark\)}}

\title{\LARGE \bf
Geometric Retargeting: \\ A Principled, Ultrafast Neural Hand Retargeting Algorithm
}

\author{Zhao-Heng Yin$^{1,2}$, Changhao Wang$^{2}$, Luis Pineda$^{2}$, Krishna Bodduluri$^{2}$, \\ Tingfan Wu$^{2}$, Pieter Abbeel$^{1}$, Mustafa Mukadam$^{2}$% <-this % stops a space
\thanks{*This work was partially done during Z.H. Yin's intern at Meta.}% <-this % stops a space
\thanks{*Project website: \href{https://zhaohengyin.github.io/geort}{zhaohengyin.github.io/geort}}%
\thanks{*For application in DexterityGen: \href{https://zhaohengyin.github.io/dexteritygen}{zhaohengyin.github.io/dexteritygen}}%
\thanks{$^{1}$BAIR, UC Berkeley EECS.}%
\thanks{$^{2}$FAIR at Meta.}%
}
\begin{document}

\maketitle
\thispagestyle{empty}
\pagestyle{empty}

%%%%%%%%%%%%%%%%%%%%%%%%%%%%%%%%%%%%%%%%%%%%%%%%%%%%%%%%%%%%%%%%%%%%%%%%%%%%%%%%
\begin{abstract}
We introduce Geometric Retargeting (GeoRT), an ultrafast, and principled neural hand retargeting algorithm for teleoperation, developed as part of our recent Dexterity Gen (DexGen) system~\cite{yin2025dexteritygen}. GeoRT converts human finger keypoints to robot hand keypoints at 1KHz, achieving state-of-the-art speed and accuracy with significantly fewer hyperparameters. This high-speed capability enables flexible postprocessing, such as leveraging a foundational controller for action correction like DexGen. GeoRT is trained in an unsupervised manner, eliminating the need for manual annotation of hand pairs. The core of GeoRT lies in novel geometric objective functions that capture the essence of retargeting: preserving motion fidelity, ensuring configuration space (C-space) coverage, maintaining uniform response through high flatness, pinch correspondence and preventing self-collisions. This approach is free from intensive test-time optimization, offering a more scalable and practical solution for real-time hand retargeting.
\end{abstract}

%%%%%%%%%%%%%%%%%%%%%%%%%%%%%%%%%%%%%%%%%%%%%%%%%%%%%%%%%%%%%%%%%%%%%%%%%%%%%%%%

\section{INTRODUCTION}
Teleoperation is essential for collecting robotic manipulation data, as it allows humans to remotely control robots in real-time. In dexterous manipulation, a fundamental component of teleoperation is kinematic retargeting~\cite{handa2020dexpilot,qin2023anyteleop,ding2024bunny,cheng2024open,wang2024dexcap}. It involves translating human gestures into corresponding robot hand poses, enabling intuitive control of robotic systems. However, defining an effective kinematic retargeting function remains a longstanding challenge. The complexity arises from the need to account for variations in human and robot configurations, and the desired level of precision. Despite the progress in this area, no universal methods have been developed that reliably captures human intent while maintaining natural and efficient robot motion. 
\begin{figure}[t]
    \centering
    \includegraphics[width=\linewidth]{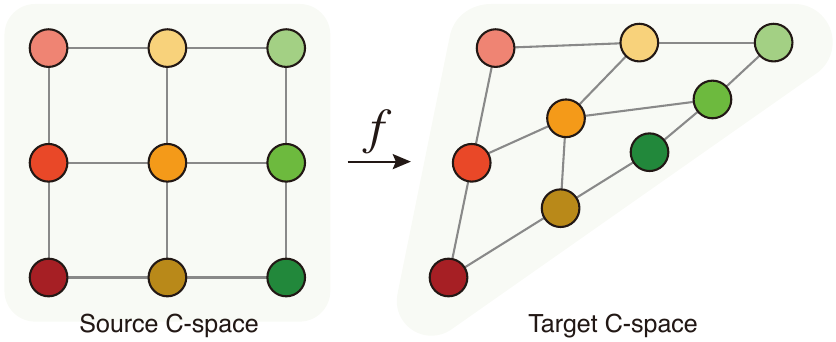} 
    \caption{Retargeting is an unconstrained problem. There are many valid retargeting functions (e.g. by dragging the point anchors in the figure). However, it is unclear how to define a proper cost functional~(objective) to specify desired retargeting function. }
    \label{fig:retargeting}
\end{figure}

\begin{figure*}[t]
    \centering
    \includegraphics[width=\textwidth]{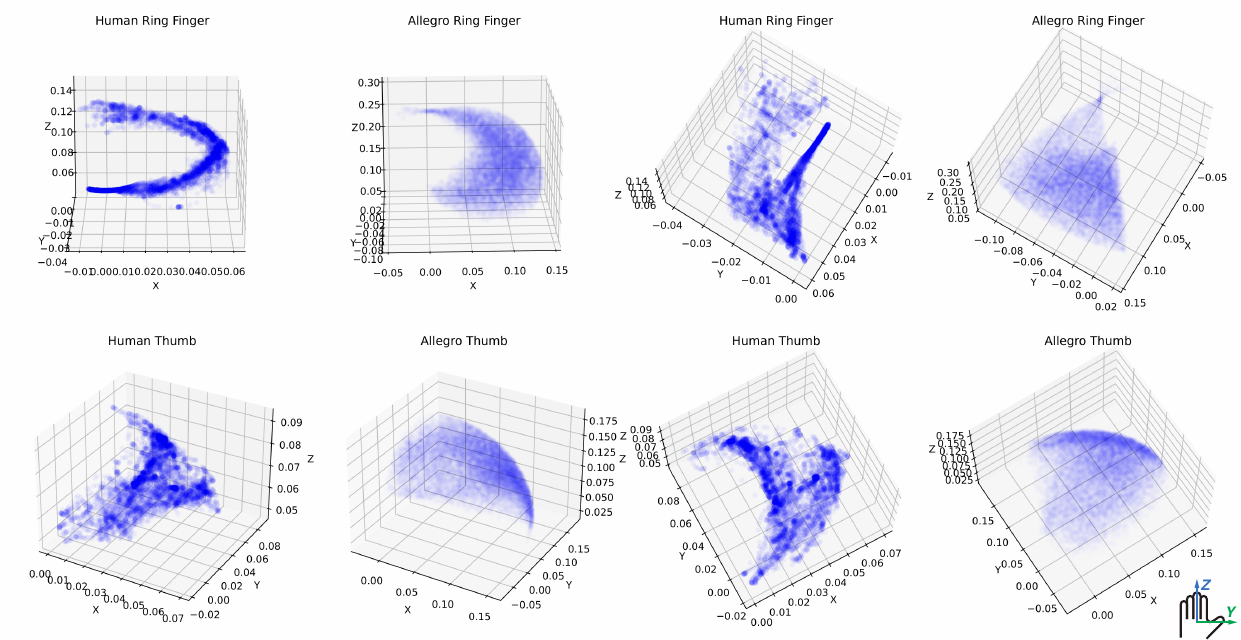} 
    \caption{Nonlinear Nature of Retargeting: In this figure, we compare shapes of human and robot~(Allegro) fingertip keypoint $C$-space (i.e. the moving range of fingertip in the hand frame). The top row shows the ring finger keypoint space comparisons and the bottom row shows the thumb keypoint space comparisons. We find that the robot and human hands keypoint spaces are not directly related through a linear mapping as suggested by previous works. In this paper, we propose novel objectives to overcome this limitation. In this figure, the robot finger keypoints are produced by random sampling in joint space and then computing forward kinematics. The human finger keypoints are produced by motion capture of a 5-minute play, in which the human is asked to move their fingers randomly to explore the limit of their hand joints.}
    \label{fig:compare}
\end{figure*}
One of the main challenges is determining the criteria for effective kinematic retargeting. Although there are numerous possible mappings from a human hand to a robot hand, the simplest and most effective criteria (objective function) for specifying and training desirable retargeting functions remain unclear. Existing methods~\cite{handa2020dexpilot,qin2023anyteleop,naughton2024respilot,sivakumar2022robotic} typically rely on a complex set of task vector constraints that ensures the retargeted robot hand pose visually looks similar to the original human hand pose. Most recent teleoperation works~\cite{sivakumar2022robotic} typically take the following linear matching form in their pipeline:
\begin{equation}
    \mathcal{L} = \sum_{i=1}^N \Vert \alpha_i \textbf{v}_H^i - \textbf{v}_R^i\Vert^2+ \text{Regularizer}.
\end{equation}
Here $\textbf{v}_H^i$ and $\textbf{v}_R^i=f(\textbf{v}_H^i)$ are the task vectors of (source) human and (retargeted) robot hands, $\alpha_i$ are some scaling hyperparameter, and $N\approx 10$ is the number of hand keypoints. The second regularization term is usually used to ensure smoothness. This formulation has several drawbacks. First, it requires several hyperparameters (e.g. $\alpha_i$, task vector origin $o_i$) to recenter and rescale each human keypoint (or task vectors), which are difficult to specify and vary between individuals. It requires a tedious process to calibrate these task-vector-related hyperparameters. Second, we notice that this linear matching objective may also be suboptimal. To illustrate this, we use Allegro Hand as an example to compare the shape of human and robot fingertip keypoint space, as shown in Figure~\ref{fig:compare}. We observe that the human fingertip keypoint $C$-space~(moving range) typically has a more curved and narrower shape, while that of the robot hand is more regular and wider. Consequently, the linear matching objective can fail to capture the correspondence and we need a more principled way to define the retargeting objective.

\begin{table}[t]
    \centering
    \renewcommand\arraystretch{1.2}
    \setlength\tabcolsep{1.7pt}
    \captionof{table}{Comparison of technical specifications of existing approaches. The speed is claimed by the referred paper or its follow up work. }
    \begin{tabular}{lcccc}
        \toprule
        Method & DexPilot~\cite{handa2020dexpilot} & AnyTeleop~\cite{qin2023anyteleop} & RTelekinesis~\cite{sivakumar2022robotic} & Ours \\
        \hline
        Hyperparams & $\geq 10$ & $\geq 10$ & $\geq 10$ & $\leq 5$\\         
        No Task Vector & \redcross & \redcross & \redcross & \greentick \\
        No Online Opt. & \redcross &  \redcross & \greentick & \greentick \\
        Retargeting Speed & 60-100Hz &  60-100Hz & 1000Hz & 1000Hz \\
        \bottomrule
    \end{tabular}
    
    \label{tab:spec}
\end{table}

\begin{figure}[h]
    \centering
    \includegraphics[width=\linewidth]{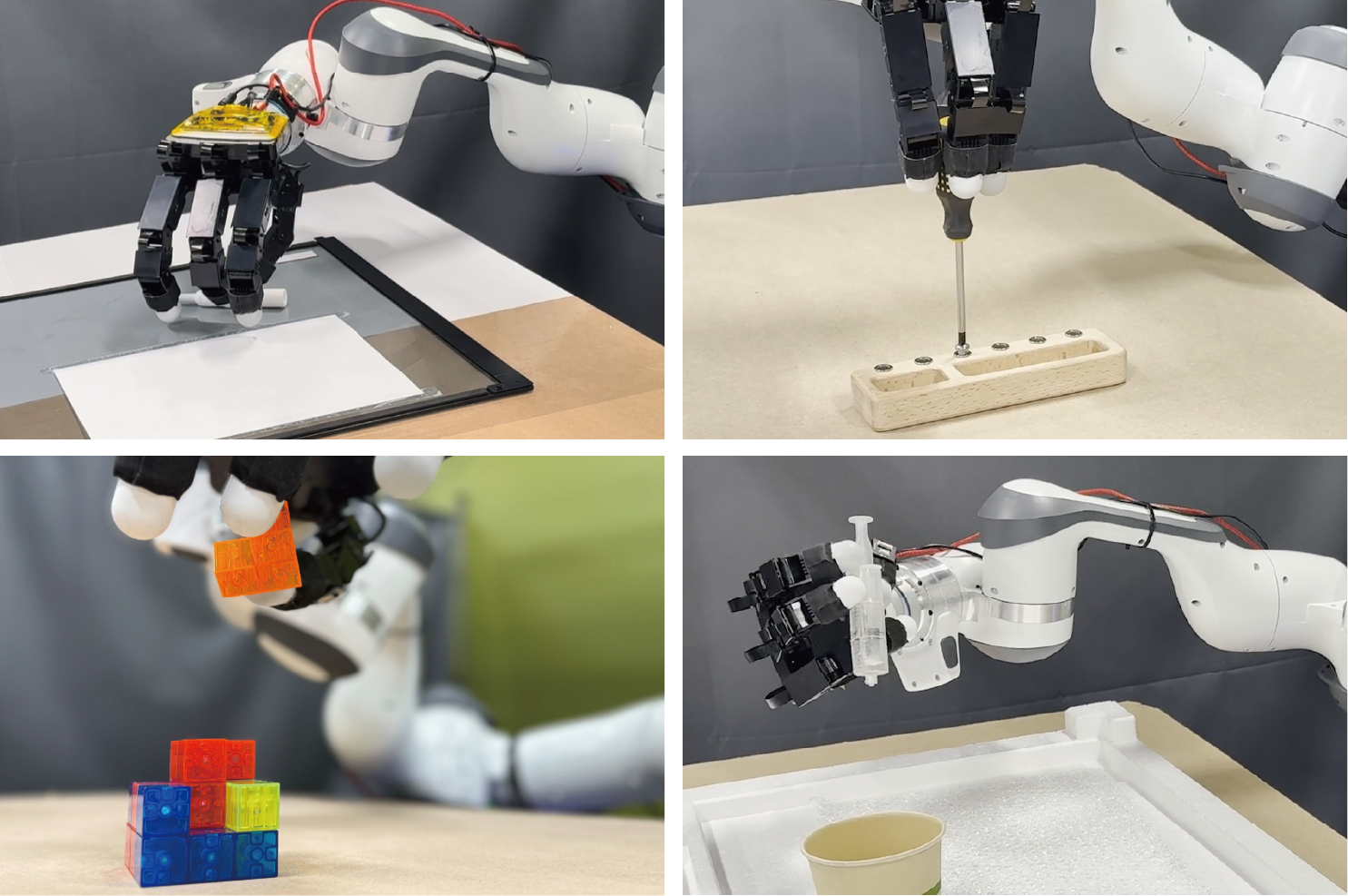} 
    \caption{The proposed principled and ultrafast teleoperation algorithm enables large-scale foundation controllers, unlocking the potential for more dexterous teleoperation systems like DexterityGen~\cite{yin2025dexteritygen}.}
    \label{fig:application}
\end{figure}

In this paper, we propose Geometric Retargeting~(GeoRT), a principled retargeting objective and training pipeline. Since the ultimate goal of retargeting is to give the human operator a sense of intuitive control over the robot hand, we define the ideal retargeting through a set of straightforward geometric criteria that characterize such requirements. We suggest that the retargeting model should (1) preserve human motion locally and preserve pinch grasps, (2) maximize $C$-space coverage so that the robot hand is fully utilized, and it should be (3) flat for uniform control sensitivity , (4) preserve pinch correspondence, and (5) collision-free. These objectives are simple to implement while providing a principled specification of retargeting quality. We also show that these principles are independent and they form minimal constraints for defining retargeting. We compare the technical specifications of our method to existing approaches in Table~\ref{tab:spec}. Our method has fewer hyperparameters and does not use heuristic task vectors, while still achieving state-of-the-art inference speed. In the experiments, we show that our algorithm has much better hand utilization and achieves better smoothness. It also outperforms existing methods in the teleoperation-based grasping task in real world experiments.

In summary,  this paper makes the following contributions: (1) We propose principled retargeting objectives for learning neural retargeting models. (2) We develop a fast neural retargeting system based on these objectives, which outperforms existing approaches in both retargeting quality and teleoperation performance and supports further applications such as DexterityGen~\cite{yin2025dexteritygen}.

\section{Geometric Retargeting}

\begin{figure*}[t]
    \centering
    \includegraphics[width=1\linewidth]{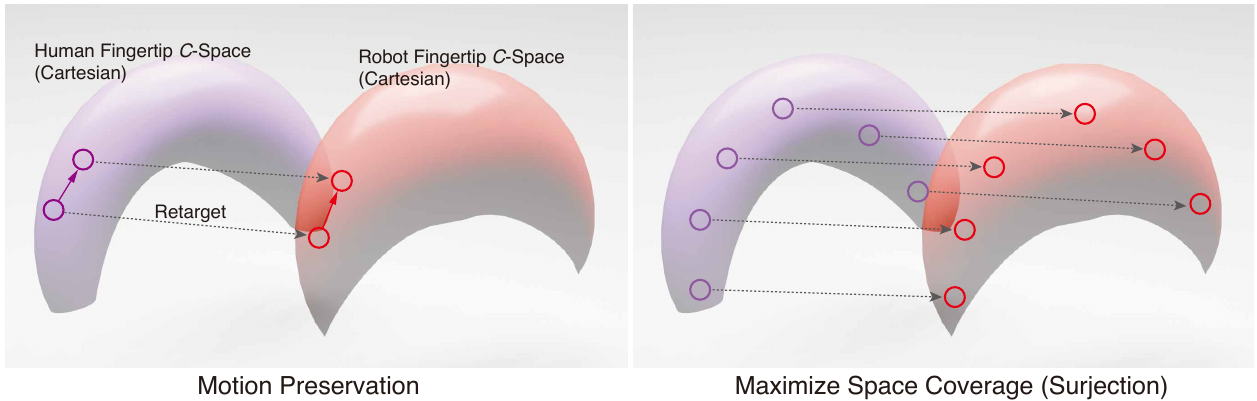}
    \caption{Basic idea of our geometric objective functions (criterion I and II). (Left) A good retargeting function should preserve the moving direction of the fingertip. (Right) Besides, the retargeting function should also be a surjection, so that the robot fingertip C-space is fully utilized. Note that we only show the $C$-space for one fingertip (e.g. index finger) in the figure.}
    \label{fig:enter-label}
\end{figure*}
\subsection{Preliminaries}
We make the following commonly used assumption as previous works. \textbf{A1}. First, we assume that the robot hand is anthropomorphic so that a natural and intuitive retargeting function may exist. \textbf{A2}. We further assume the existence of finger correspondence: e.g. humans use their index fingertip to control the robot's ``index'' fingertip. We denote the humans' and robot's fingertip position in their own wrist frame as $x_H^i$ and $x_R^i$ respectively, where $i$ is the fingertip index. 

In this paper, a kinematic retargeting model $f$ is a function that maps a set of human fingertip keypoints to robot hand joint positions, which is different from works that also take the object model and poses as input for joint hand-object retargeting. 

\subsection{Criterion I: Motion Preservation}
We require the retargeting function to preserve the movement direction of each fingertip. This aligns with a fundamental expectation in teleoperation: when a human operator moves their finger in a certain direction, they naturally expect the robot’s fingertip to follow the same trajectory. Formally, given any position $x_H^i$ for $i$-th finger and any small moving direction $d$, we require $d$ parallel to $\text{FK}_i \circ f_i(x_H^i+d) - \text{FK}_i \circ f_i(x_H^i)$, where $\text{FK}_i$ is the forward kinematics for fingertip $i$, and $f_i$ is the retargeting component for $i$-th finger. This criterion can be described by a simple loss function:
\begin{align}
    \mathcal{L}_{dir} &= - \sum_{i}\mathbb{E}_{d,x_H^i} \text {Dir}(x_H^i, d) \\ &= - \sum_i\mathbb{E}_{d,x_H^i} \langle \frac{d}{\Vert d\Vert}, \frac{\text{FK}_i \circ f_i(x_H^i+d) - \text{FK}_i \circ  f_i(x_H^i)}{\Vert \text{FK}_i \circ  f_i(x_H^i+d) - \text{FK}_i \circ f_i(x_H^i)\Vert}\rangle.
\end{align}
We implement $\text{FK}_i$ as a pretrained neural forward kinematics function. One can also use an analytical forward kinematics function.

\begin{figure}[h]
    \centering
    \includegraphics[width=\linewidth]{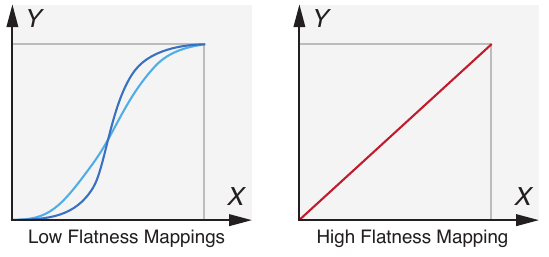} 
    \caption{The retargeting mapping should have a high flatness (criterion III). In this 1D retargeting example~(mapping an interval on the $x$-axis to another interval on the $y$-axis), this is equivalent to $f'(x)$ being constant everywhere, so that any small $\Delta x$ will lead to the same amount of $\Delta y$. Note that the blue curves on the left can satisfy the criterion I and II. Therefore, introducing a third flatness objective is necessary.} 
    \label{fig:flatness}
\end{figure}

\subsection{Criterion II: $C$-space Coverage}
Besides motion preservation, another important criterion is $C$-space coverage. Intuitively, we want the robot's $C$-space to be fully utilized—when a human moves their fingers from one limit to the other, the robot hand should replicate this motion across its full range, rather than being confined to a limited subset of its $C-$space. 

Formally, we denote the keypoint $C-$space of $i$-th robot and human fingertip as $KC^i_R$ and $KC^i_H$ respectively. Then, our coverage criteria states that $\text{FK}_i\circ f_i$ should be a surjection from $KC^i_H$ to $KC^i_R$, and we should minimize the volume of $KC^i_R \setminus (\text{FK}_i\circ f_i(KC^i_H))$, i.e. the uncovered $i$-th fingertip keypoint $C$-space of robot hand.

However, computing this uncovered space and its volume is computationally expensive, and this does not yield a differentiable function either. Therefore, we propose to use Chamfer loss~\cite{barrow1977parametric} in 3D vision research as a proxy for this procedure. In each minibatch, we sample $P_H^i\sim KC^i_H$ and $P_R^i\sim KC^i_R$ uniformly, and we minimize 
\begin{equation}
    \mathcal{L}_{cover} = \sum_i\mathbb{E}_{P_H^i\sim KC^i_H,P_R\sim KC^i_R}\text{Chamfer}(P_R^i, \text{FK}_i\circ f_i(P_H^i)).
\end{equation}
This loss guarantees that any random point cloud representation of $ KC^i_R$ can be closely approximated by projecting the corresponding representation of $ KC^i_H$ with retargeting model.

\subsection{Criterion III: High Flatness}
While the last two criteria already define a reasonable retargeting mapping, we find it necessary to introduce a flatness objective to ensure that the model responds uniformly as the user moves across the $C$-space, which enhances the predictability and intuitiveness of the interaction experience. 
To understand this, we illustrate an 1D example in Figure~\ref{fig:flatness}. The retargeting functions defined by the blue curves satisfy the criterion I and II simultaneously, however, the same $\Delta x$ in the source $X$ space may lead to different $\Delta y$ in the target $Y$ space. In this case, the user may perceive the retargeting as too unresponsive~(i.e. $f'(x)\approx 0)$ in some areas, while overly sensitive~(i.e. $|f'(x)|$ too large) in others. The ideal retargeting should have high flatness, which means $f'(x)$ being almost constant, or a low $f''(x)=\frac{d^2f}{dx^2}(x)\approx 0$ equivalently. To generalize this idea to high dimensional space, we propose to minimize $\mathbb{E}_{x_H, d} \left\Vert\frac{d^2(\text{FK}\circ f)}{dt^2} (x_H+td)\right\Vert^2$, which means the second-order directional derivative at any point along any direction should be close to 0. We use the finite difference method to evaluate the derivatives, yielding the following objective function:
\begin{equation}
    \mathcal{L}_{flat} =  \mathbb{E}_{x, d} \Vert \text{FK}\circ f(x+d) + \text{FK}\circ f(x-d)-2\text{FK}\circ f(x)\Vert^2.
\end{equation}

The global linear matching objective (i.e. Equation 1) used by previous works naturally encourages flatness. However, for general non-linear retargeting problems, perhaps the best way is to use the proposed local flatness constraint.

Note that criteria II and III can not derive criterion I (motion preservation). In the example, reversing the linear mapping (change the slope to its opposite and shift) on the right can still make criteria II and III hold, but the moving direction in the target space will be reverted and very unintuitive. 

\begin{figure}[t]
    \centering
    \includegraphics[width=1\linewidth]{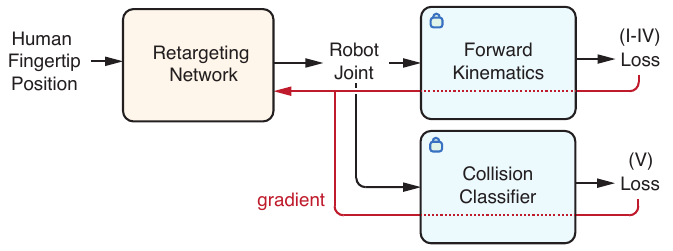}
    \caption{Model training update procedure. The geometrical loss functions~(I-IV) are computed in the keypoint spaces~(after the differentiable forward kinematics). The gradient backpropagates through the forward kinematics model and the collision classifier to the retargeting network. Note that the forward kinematics model and collision classifier are only used during training.}
    \label{fig:enter-label}
\end{figure}
\subsection{Criterion IV: Pinch Correspondence}
The previous criterion focused more on per-finger motion regulation.
Another important expectation from users is pinch correspondence. For example, when the users do a pinch grasp using the thumb and index finger, they typically expect that the robot hand does the same. We find this crucial to provide users with a sense of agency, however, previous criteria do not strictly guarantee this and sometimes we notice that the emerged pinch correspondence is not perfect. To improve this, we further introduce a pinch correspondence constraint. Specifically, for a human gesture $x_H$, if $x_H^i - x_H^j$ is below a threshold $d$ (such as 1cm), i.e., finger $i$ and $j$ are pinching, then we require $\text{FK}_i\circ f_i(x_H^i)$ close to $\text{FK}_j\circ f_j(x_H^j)$. This can be written as
\begin{align}
    &\mathcal{L}_{pinch} = \mathbb{E}_{x_H}  \\
    &\sum_{(i,j):i\neq j} \mathbf{1} (\Vert x_H^i - x_H^j\Vert< d)\Vert \text{FK}_i\circ f_i(x_H^i) - \text{FK}_j\circ f_j(x_H^j)\Vert^2.
\end{align}
Note that this requires human users to provide some pinch grasp examples. Fortunately, this can be easily collected within 5 minutes of random play as we will discuss in the implementation section.

\subsection{Criterion V: Collision-Free Retargeting}
Finally, a collision-free human hand gesture should correspond to a collision-free robot hand gesture. Therefore, we introduce collision-free as our final criterion. Similar to~\cite{sivakumar2022robotic}, we first pretrain a collision classifier $C$ to decide the probability of a joint configuration $q$ leading to hand self-collision. The training dataset is generated through simulation, and the label is obtained by querying a collision checker to get the binary self-collision label. Then, we use the following collision loss over retargeting model $f$:
\begin{equation}
    \mathcal{L}_{col} = -\mathbb{E}_{x_H}\log (1-C(f(x_H)));
\end{equation}
Note that $C$ is fixed as we train $f$. Interestingly, we find that even without this term our loss can lead to few collisions for certain robot hands. Nevertheless, we introduce this for completeness.

\begin{algorithm}[t]
\caption{Geometric Retargeting}\label{alg:geort}
\begin{algorithmic}[1]
\STATE Generate (joint position $q$, fingertip position $x$, collision $c$) data in simulation to train each neural forward kinematics models $\text{FK}_i$ and collision classifier $C$.
\STATE Generate point cloud approximation $\widetilde{KC}_R^i, \widetilde{KC}_H^i$ of each keypoint $C$-space $KC_R^i, KC_H^i$. 
\FOR{$\text{itr}=0,1,..., N_{train}$}
    \STATE Sample $\textbf{d}\sim \mathcal{N}(0, \sigma^2)$ and random human gesture $\textbf{x}_H$ to compute $\mathcal{L}_{dir},\mathcal{L}_{flat},\mathcal{L}_{pinch}, \mathcal{L}_{col}$; 
    \STATE Sample $\textbf{P}_H^i\sim \widetilde{KC}_H^i$,  $\textbf{P}_R^i\sim \widetilde{KC}_R^i$ to compute $\mathcal{L}_{cover} $;
    \STATE Optimize $f$ by taking gradient descent with $\mathcal{L} = \mathcal{L}_{dir}+\lambda_1\mathcal{L}_{cover} + \lambda_2\mathcal{L}_{flat}+\lambda_3\mathcal{L}_{pinch}+\lambda_4\mathcal{L}_{col}$.
\ENDFOR
\RETURN $f$
\end{algorithmic}
\end{algorithm}

\subsection{Implementation}
We implement our kinematic retargeting model as a set of independent retargeting models. For example, for the Allegro Hand which has four fingers, we define $f(x_H^1, x_H^2, x_H^3, x_H^4) = [f_1(x_H^1), f_2(x_H^2), f_3(x_H^3), f_4(x_H^4)]$, and each $f_i$ is an independent retargeting model for the corresponding finger. We parameterize each $f_i$ as a multi-layer perception~(MLP). We also rescale the joint position range to $[-1,1]$ and use Tanh as output activation of each $f_i$. We use a combination of the proposed loss functions to train our model:
\begin{equation}
    \mathcal{L} = \mathcal{L}_{dir}+\lambda_1\mathcal{L}_{cover} + \lambda_2\mathcal{L}_{flat}+\lambda_3\mathcal{L}_{pinch}+\lambda_4\mathcal{L}_{col}.
\end{equation}

This optimization objective only has 4 hyperparameters compared to previous works that have numerous scale hyperparameters and heuristic task vectors. We present our training loop in Algorithm~\ref{alg:geort} for clarity. The training procedure is very fast in practice, only taking 3-5 minutes on a single NVIDIA 3060 GPU. The point cloud approximation of the robot hand fingertip keypoint $C$-space can be generated by randomly moving the hand in simulation. For that of human hand, we find the following method effective: we ask the human user to stretch their fingers and move back and forth, as well as perform various pinch grasps under any motion capture system (or other hand tracking system such as a glove). We record the moving trajectory of each fingertip keypoint, giving us several raw point clouds. This data collection process is fast and typically takes less than 5 minutes and is a one-time calibration. We find that an empirical loss weight setup that works well is $\lambda_1\in[10,100], \lambda_2=1, \lambda_3\in[10^3, 10^6], \lambda_4=[10^{-4}, 10^{-2}]$. Varying the weight inside these intervals can change the retargeting details a bit, but overall they look similar and provide good results. Note that the magnitudes of some weights are large due to the distance unit we use.

\begin{figure}[t]
    \centering
    \includegraphics[width=1\linewidth]{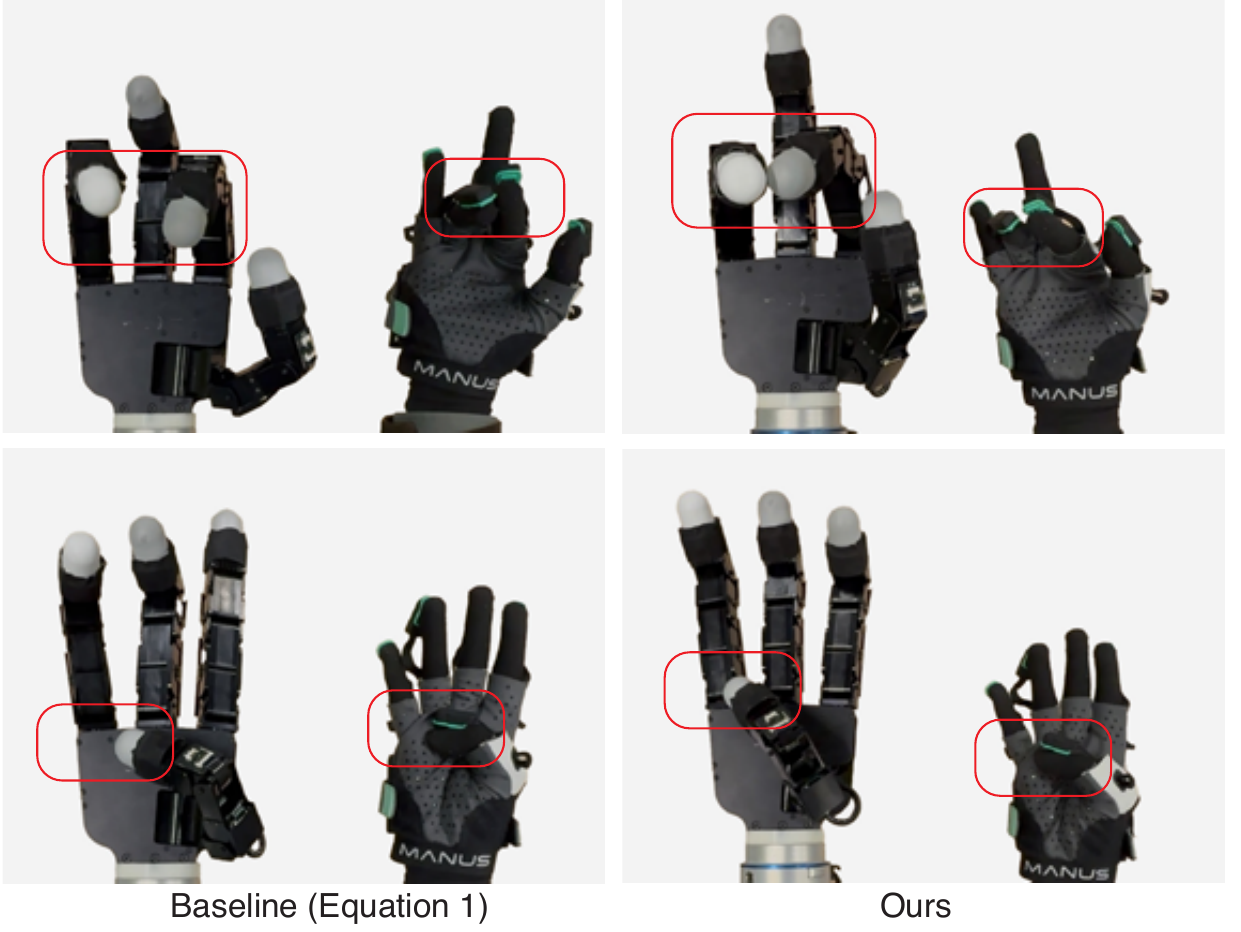}
    \caption{Qualitative comparisons. We find that due to insufficient $C$-space coverage, the baseline method fails to provide important functionalities such as index-ring finger pinch, which is essential for in-hand manipulation. }
    \label{fig:qual_comp}
\end{figure}
\section{Experiments}
In this section, we compare the grasping performance of this work to other teleoperation methods. We also study the properties and design choices of our neural retargeting method. For its application, we refer readers to the DexterityGen paper.

\begin{table}[t]
    \centering
    \normalsize
    \setlength\tabcolsep{1.7pt}
    \renewcommand\arraystretch{1.2}
    \captionof{table}{Quality metric of different loss objectives. Note that the online version of Eqn (1) is not time-independent and its result relies on the previous frame, so we use its offline version as an approximation. }
    \begin{tabular*}{\linewidth}{l@{\extracolsep{\fill}}ccc}
        \toprule
           \textbf{Method} & \multicolumn{2}{c}{\textbf{Equation 1}}  & \textbf{Ours} \\
              & Offline~\cite{sivakumar2022robotic} & Online~\cite{handa2020dexpilot, qin2023anyteleop} & \\
        \hline
        Motion Preservation$(\uparrow)$ & 0.73 & -- & \textbf{0.94} \\
        $C$-space coverage$(\uparrow)$ & 38\% & -- & \textbf{90\%} \\
        \bottomrule
    \end{tabular*}
    \label{tab:metric}
\end{table}
\begin{table}[t]
    \centering
    \normalsize
    \renewcommand\arraystretch{1.2}
    \captionof{table}{Teleoperation performance comparison of different methods in real world. Our method offers a faster and more effective teleoperation experience.}
   \begin{tabular*}{\linewidth}{l@{\extracolsep{\fill}}ccc}
        \toprule
        \textbf{Method} & \multicolumn{2}{c}{\textbf{Equation 1}}  & \textbf{Ours} \\
              & Offline~\cite{sivakumar2022robotic} & Online~\cite{handa2020dexpilot, qin2023anyteleop} & \\
        \hline
        Onetime-Success$(\uparrow)$ & 55\% & 42.5\% & \textbf{87.5\%} \\
        Completion Time$(\downarrow)$ & 9.0s & 19.3s & \textbf{3.2s} \\
        \bottomrule
    \end{tabular*}
    \label{tab:grasp}
\end{table}

\begin{figure*}[t]
    \centering
    \includegraphics[width=1\linewidth]{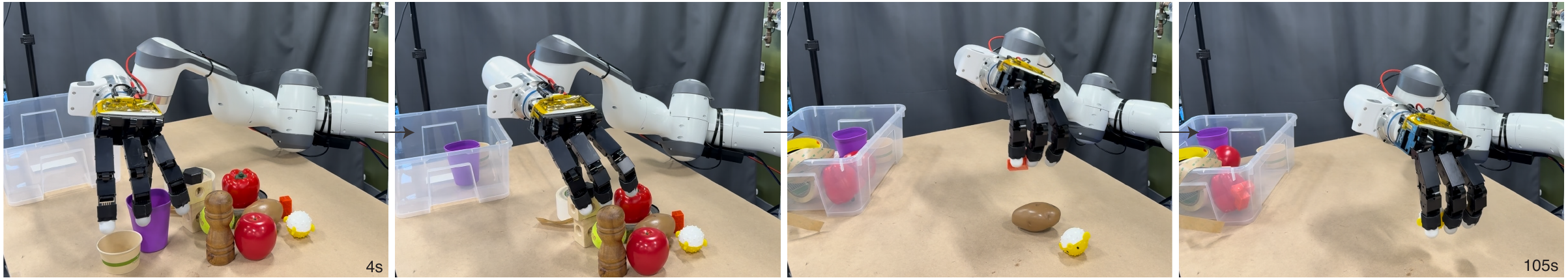}
    \caption{The user can use our system to clean up a pile of objects~(12) on the table in around 100 seconds easily. Note that the relatively slow arm motion is the main bottleneck here. The user can grasp most of the objects successfully with 1 trial.}
    \label{fig:enter-label}
\end{figure*}
\begin{figure*}[t]
    \centering
    \includegraphics[width=1\linewidth]{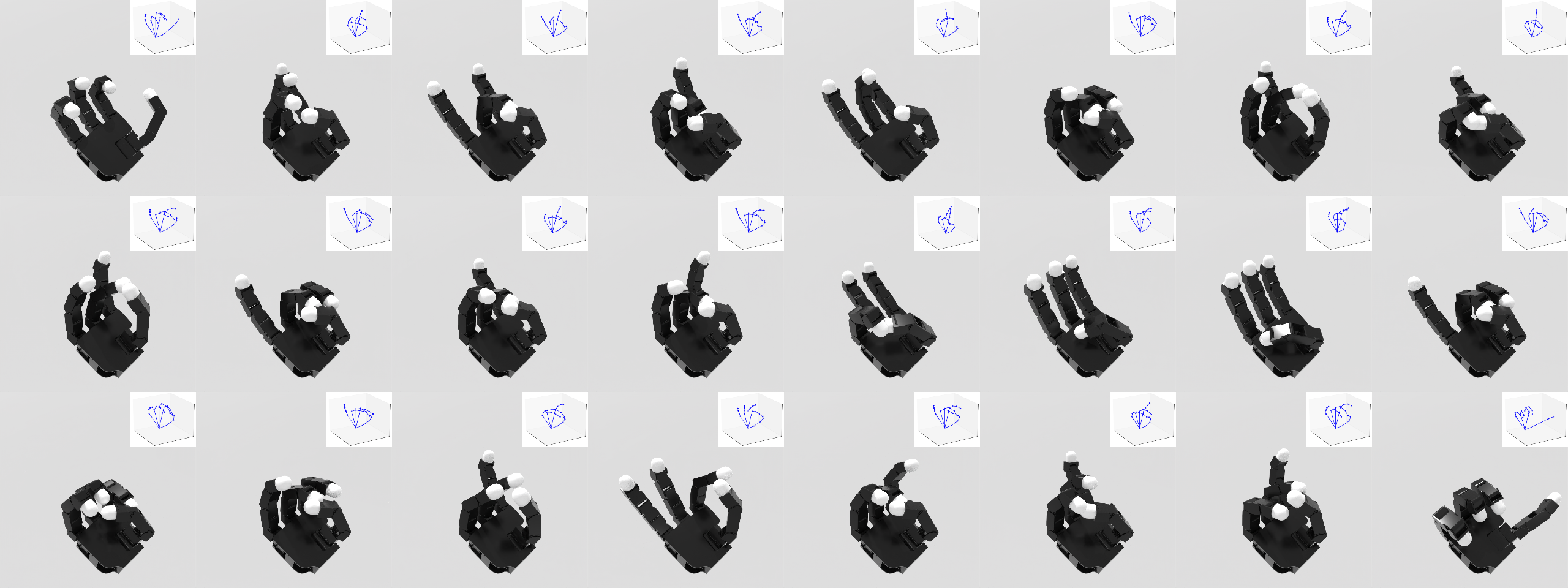} \\
    \vspace{1mm} % Adds vertical space
    \includegraphics[width=1\linewidth]{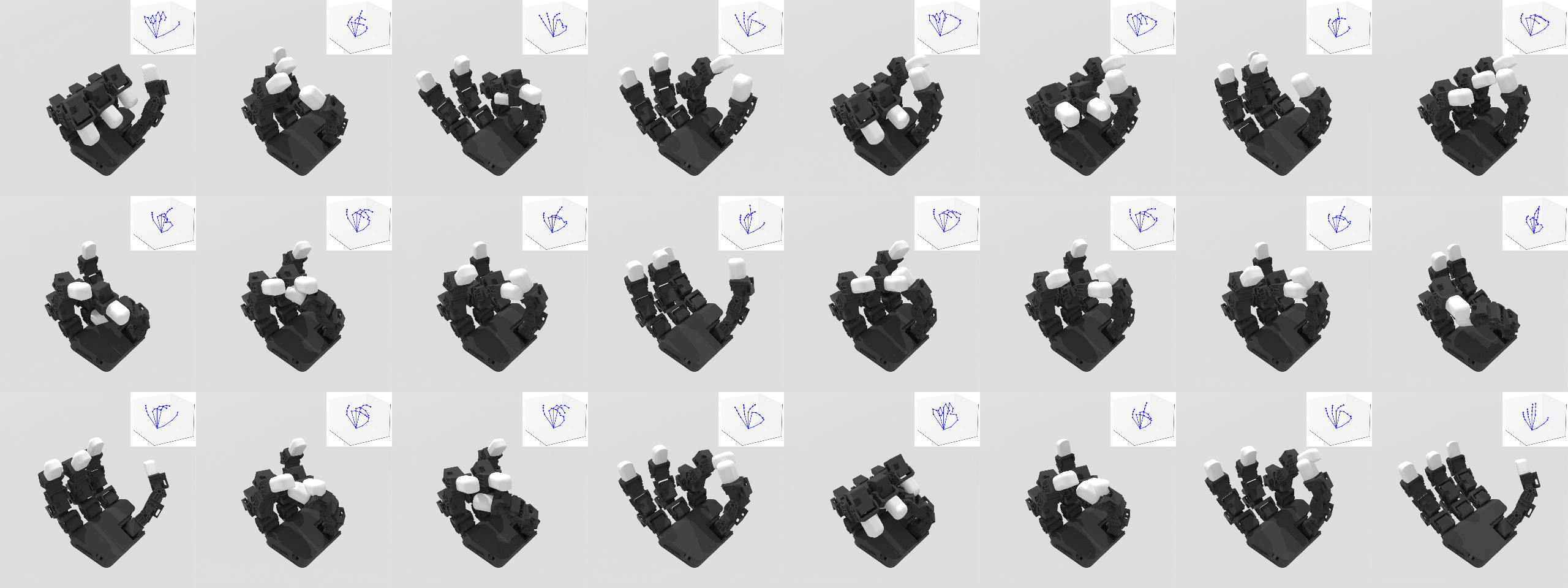}
    
    \caption{Qualitative retargeting results of our retargeting method. The top half is the Allegro hand and the bottom half is the Leap hand. Even if our method does not use any task-vector-based matching terms, it can discover the correspondence between human and robot hands.}
    \label{fig:qual}
\end{figure*}

\subsection{Simulation Evaluation}
We first compare the quality of different methods in simulation. We quantitatively measure the smoothness score and the $C$-space coverage score. The definitions of these two metrics are as follows:
\begin{enumerate}
    \item \textbf{Motion Preservation} is related to Criteria I. We compute this by uniformly sample anchor points $x_H^i$ and directions $d$, and we compute $\frac{1}{N}\sum_{i=1}^N\mathbb{E}_{d,x_H^i} \text {Dir}(x_H^i, d)$, i.e. how good the robot hand movement is aligned with human hand movement. The metric is bounded by $[-1, 1]$.
    \item \textbf{$C$-space Coverage} is related to Criteria II, quantifying the effective moving range of the fingertip. However, we do not exactly compute this following the proposed objective function. We sample sufficiently many points $x_H^i$ in each human fingertip keypoint $C$-space and we compute how much they occupy the robot keypoint $C$-space, given by $\text{Vol} (\cup_i\mathcal{B}(\text{FK}\circ f(x_H^i), r)\cap KP_R^i)/\text{Vol}(KP_R^i)$. Here, $\mathcal{B}(x,r)$ denotes a sphere or radius $r$ centered at $x$. This metric is between $[0\%, 100\%]$.
\end{enumerate}
\textbf{Results}
We list the evaluation results of different methods on Allegro Hand in Table~\ref{tab:metric}. We find that our approach can achieve much better smoothness and $C$-space coverage compared to baseline methods. However, this result is expected since we directly use these two metrics for optimization. This suggests that our approach can fully utilize the moving range of the robot hand while providing the user with a smooth sense of control. \ \\ \\
\textbf{Qualitative Results} Furthermore, we investigate whether our quantitative results translate to plausible retargeting. We plot some retargeting results in Figure~\ref{fig:qual_comp}. Surprisingly, even if we do not use any task vector matching heuristics, a good correspondence emerges from our simple objective function on different hands~(Allegro and LEAP~\cite{shaw2023leap} hand). We also compare our method to the baseline in Figure~\ref{fig:compare}. We find that due to insufficient coverage, the baseline method fails to utilize the lateral movement of fingers effectively and cannot do effective pinch grasp or finger~(thumb) reaching.

\subsection{Real-world Experimental Setup}
Then, we evaluate our approach on an arm-hand robotic system. In this paper, we use the Allegro robot hand with the Franka Panda robot arm. For the teleoperation of Allegro hand, we first use a Manus glove to capture the human hand keypoints. These keypoints are then fed into the retargeting model to produce the allegro hand joint target. These joint target are then sent to a PD controller to drive the hand. For the arm teleoperation, we use a Vive tracker system to capture the human wrist pose and use it to control the motion of robot arm's end-effector pose.

\subsection{Real-world Evaluation}
Since performing dexterous in-hand manipulation is hard as suggested by previous works, in this paper we mainly consider the grasping performance, which is also a crucial step in robotic manipulation. We record the one-trial success rate and the average time a user takes to grasp an object successfully.  \ \\
\textbf{Results} We show the result in Table \ref{tab:grasp}. We find that our method can allow for faster and more effective grasping in real world. We account for this by better utilization of fingertip $C$-space and smoother and more intuitive retargeting. Specifically, we find that it is hard to grasp tiny objects with the baseline method due to the unintuitive finegrained control of fingertip.

\section{Related Works}
\textbf{Retargeting}
Retargeting is an important step in teleoperation. Some works propose to use joint-space retargeting~\cite{liu2017glove, rajeswaran2017learning}, which maps the joint of the human hand to that of the robot hand through some predefined mapping. Although this approach is intuitive in some cases, it fails to provide precise control in general due to differences in kinematic structure between the robot hand and the human hand. Some works also propose direct cartesian mapping~\cite{handa2020dexpilot} from human hand keypoint to robot hand keypoint and use an inverse kinematics model to decide the hand joints. Most of the recent works in robot hand teleoperation apply this cartesian keypoint mapping approach. However, specifying keypoint mapping is nontrivial as we discussed and they can lead to unnatural hand poses. Instead of using some heuristic cartesian mapping rule~(e.g. linear rule), in this paper we propose novel objectives based on local motion and global $C$-space matching, avoiding the challenge of designing complex heuristics. In the retargeting literature, some works also consider task-oriented retargeting which takes object state as input~\cite{lakshmipathy2024kinematic}, and this setup is different from ours. However, we believe that our proposed regularization can also be used to improve these methods. We refer readers to~\cite{meattini2022human} for a comprehensive review of existing retargeting approaches.

\textbf{Shape Correspondence}
Our idea is also related to shape correspondence research in vision and 3D data processing research. The goal of shape correspondence is to find some homeomorphic mapping between two manifolds~\cite{kohonen1990self}, and the idea can also be applied to direct cartesian mapping~(retargeting). Existing work typically defines some form of energy or cost functional for the whole mapping, such as elastic energy to minimize distortion. There is a rich literature on learning from functional maps, and the proposed methods have been applied to define a correspondence between different object shapes~(e.g. human to human correspondence)~\cite{attaiki2021dpfm, pai2021fast, sharp2022diffusionnet}. However, so far this line of research has not been applied to retargeting systematically yet. A recent retargeting work~\cite{chong2021learning}, harmonic autoencoder, also leveraged ideas in this field~(e.g. using chamfer loss) to improve mapping quality. However, it also uses pairwise human-to-robot data in training. In contrast, we propose using motion and flatness losses as supervision, which makes our method fully unsupervised.

\section{Conclusion}
In this paper, we have presented Geometric Retargeting (GeoRT) a fast, efficient, and principled approach to neural hand retargeting for teleoperation, integrated into the Dexterity Gen (DexGen) system. GeoRT achieves state-of-the-art speed with minimal hyperparameters compared to existing methods. Its unsupervised training eliminates the need for manual hand pair annotations, while its novel geometric objective functions ensure both motion fidelity and C-space coverage. GeoRT's high-speed performance and scalability make it a practical and flexible solution for real-time hand retargeting without the need for intensive test-time optimization.

\section*{Acknowledgments}
This work was partially carried out during Zhao-Heng Yin's intern at the Meta FAIR Labs. This work is supported by the Meta FAIR Labs. Zhao-Heng Yin's research is supported by ONR MURI N00014-22-1-2773. Pieter Abbeel holds concurrent appointments as a Professor at UC Berkeley and as an Amazon Scholar. This paper describes work performed at UC Berkeley and is not associated with Amazon.

\bibliographystyle{unsrt}
\bibliography{reference}

\end{document}